%% file: main.tex
\documentclass[10pt,twocolumn,letterpaper]{article}

\usepackage{cvpr}              

\input{preamble}

\definecolor{cvprblue}{rgb}{0.21,0.49,0.74}
\usepackage[pagebackref,breaklinks,colorlinks,allcolors=cvprblue]{hyperref}

\def\paperID{35} 
\def\confName{CVPR}
\def\confYear{2025}

\title{VRU-CIPI: Crossing Intention Prediction at Intersections for Improving Vulnerable Road Users Safety}

\author{\makebox[0.7\textwidth][c]{Ahmed S. Abdelrahman \hfill Mohamed Abdel-Aty \hfill Quoc Dai Tran}\\
University of Central Florida, United States\\
{\tt\small \{ahmed.abdelrahman, m.aty, quocdai.tran\}@ucf.edu}}

\begin{document}
\maketitle
\input{sec/0_abstract}

\input{sec/1_intro}
\input{sec/2_relatedwork}

\input{sec/3_CIPI}
\input{sec/5_Experiment}

\input{sec/6_Conc}

{
    \small
    \bibliographystyle{ieeenat_fullname}


}


\end{document}

%% file: preamble.tex
\usepackage{comment}
\usepackage{arydshln}

%% file: sec/0_abstract.tex
\begin{abstract}

Understanding and predicting human behavior in-the-wild, particularly at urban intersections, remains crucial for enhancing interaction safety between road users. Among the most critical behaviors are crossing intentions of Vulnerable Road Users (VRUs), where misinterpretation may result in dangerous conflicts with oncoming vehicles. In this work, we propose the VRU-CIPI framework with a sequential attention-based model designed to predict VRU crossing intentions at intersections. VRU-CIPI employs Gated Recurrent Unit (GRU) to capture temporal dynamics in VRU movements, combined with a multi-head Transformer self-attention mechanism to encode contextual and spatial dependencies critical for predicting crossing direction. Evaluated on UCF-VRU dataset, our proposed achieves state-of-the-art performance with an accuracy of 96.45\% and achieving real-time inference speed reaching 33 frames per second. Furthermore, by integrating with Infrastructure-to-Vehicles (I2V) communication, our approach can proactively enhance intersection safety through timely activation of crossing signals and providing early warnings to connected vehicles, ensuring smoother and safer interactions for all road users.

\end{abstract}

%% file: sec/1_intro.tex
\section{Introduction} 
\label{sec:intro}

Understanding and predicting human behavior in dynamic urban environments represents a critical challenge at the intersection of object detection, pose estimation, and activity recognition \cite{schneemann2016context}. This capability is particularly essential for autonomous driving systems and smart traffic management, where accurately interpreting the intentions of VRUs can prevent potentially fatal collisions. Among the most challenging scenarios are urban intersections, where misinterpreting a pedestrian or cyclist's crossing intentions may result in dangerous conflicts with vehicles \cite{zhou2023pedestrian}. In the United States, traffic crashes resulted in 6,516 pedestrian fatalities in 2020, a 3.9\% increase from 6272 pedestrian fatalities in 2019, and an estimated 54,769 pedestrians were injured \cite{NHTSA_Pedestrian_Safety}. In 2020, the number of bicyclist fatalities reached 938, 26\% of which occurred at intersections. This represented a 9\% increase in bicyclist fatalities, up from 859 in 2019 \cite{lawstraffic}. Pedestrian fatalities represent 17\% of all traffic fatalities, with an increase of 53\% in 2018 compared to 2009 \cite{GHSA_Pedestrian_Fatalities_2019}. Florida ranks third in terms of pedestrian fatalities per capita in the U.S., and the Orlando-Kissimmee-Sanford area is fifth among the deadliest metropolitan regions, with 656 pedestrian deaths recorded between 2008 and 2017 \cite{america2017dangerous}.

Human behavior prediction at intersections represents a complex activity recognition problem that requires sophisticated spatio-temporal modeling of body movements, poses, and contextual factors. Significant advancements have been made in monitoring their activity \cite{rasouli2021bifold, sharma2022pedestrian, abdelrahman2025video} to improve their safety \cite{tran2025leveraging, jafari2024pedestrians}. Despite these advances, understanding vulnerable road user (VRU) behavior at intersections remains challenging, as VRUs frequently exhibit risky actions such as neglecting to activate pedestrian signals or crossing during prohibited phases \cite{van2006pedestrian, bradbury2012go}. Such behaviors substantially increase the likelihood of dangerous encounters with vehicles, leading to severe conflicts and collisions \cite{autey2012safety, buch2017incidents}. VRU violations and frequent disregard for compliance emerge as leading factors contributing to VRU conflicts at intersections. Furthermore, it is imperative not only to enhance VRU safety but also to simultaneously maintain or enhance pedestrian signal performance. Improving signal performance while ensuring VRU safety is driven by the urgent need to balance the critical aspects of urban mobility \cite{ahsan2025evaluating}, the safety of VRUs and the efficiency of vehicular traffic flow. In modern urban centers, intersections are more than mere crossroads; they are pivotal points where the safety of VRUs such as pedestrians and cyclists intersects with the smooth functioning of traffic \cite{amini2022development, ma2022mapping}. Traditional traffic signal systems often struggle with the challenge of maintaining VRU safety at intersections without impeding vehicular flow, leading to a compromise on either safety or efficiency. The aim of this research work is to provide safe and ample crossing experiences for VRUs while reducing unnecessary vehicle delays caused by false calls during the pedestrian phase. This not only enhances the safety and experience of VRUs but also contributes to a more fluid and less congested traffic system \cite{guo2012reliability, brosseau2013impact, budzynski2017pedestrian, zhang2020aprediction, haque2023modeling}.

Many studies have been addressing few of these issues; however, there are three main research gaps in these studies, which are intended to be addressed in this work, as follows:

\begin{itemize}
    \item At a typical intersection, there are multiple waiting areas that each serve two crosswalks; however, most of the current studies focus only on deciding whether the VRU will cross or not, without specifying which crosswalk it is intended to go through \cite{zhang2020bprediction, haque2023modeling}.

    \item The lack of real-time processing in many of the proposed systems that address predicting VRU crossing intention \cite{zhang2020bprediction, zhang2021pedestrian, yang2022cooperative}. Achieving real-time processing stands as a critical element in the development of a practical and intelligent system suitable for deployment at intersections.

    \item Most of the datasets used are typically addresses pedestrians only missing many other VRU types such as cyclists and e-scooter riders and misses nighttime instances and raining weather conditions.
    
\end{itemize}

In this work, we address these limitations by introducing VRU-CIPI, a novel real-time framework that leverages multiple input features including pose estimation to understand their behavior combined with temporal modeling for precise crossing direction prediction. Our approach employs GRU to capture the temporal dynamics of VRU movements, integrated with a multi-head self-attention Transformer mechanism to model the spatial relationships between pose keypoints and contextual information. Deploying VRU-CIPI at intersections can aid in proactively activate corresponding pedestrian phases based on the predicted crossing directions improving road users safety.

%% file: sec/2_relatedwork.tex
\section{Related Work}
\label{sec:relatedW}

\subsection{Crossing Intention Prediction from Vehicle Camera View}

Pedestrian crossing intention prediction is crucial for safe navigation of autonomous vehicles \cite{abdelrahman2023development, abdelrahman2023scalable}, prompting extensive research using diverse deep learning architectures. Current models primarily fall into CNN-based, RNN-based, GCN-based, transformer-based, and fusion-based methods, each uniquely leveraging different modalities, features, and temporal-spatial dynamics. The most widely used datasets for onboard vehicle camera-based pedestrian intention prediction are JAAD \cite{rasouli2017they} and PIE \cite{rasouli2019pie}. Early pedestrian intention prediction models primarily relied on only convolutional neural network (CNN) to extract visual features from image data. \cite{8265243} employed VGG16 and ResNet50 backbones to process pedestrian image crops and global scene images, integrating high-level visual features with pedestrian intent labels. However, CNN-based approaches largely overlooked temporal dependencies, limiting their ability to model sequential motion patterns necessary for accurate crossing intention prediction. To address temporal modeling, recurrent neural network (RNN), including LSTM and GRU, were introduced. \cite{9304591} incorporated multi-modal features such as bounding box coordinates, vehicle speed, and pedestrian intention data to enhance prediction performance. \cite{9743954} extended this by integrating segmentation maps, improving environmental awareness. However, these models faced computational inefficiencies and struggled to capture geometric relationships within pedestrian movement. PIP-Net \cite{azarmi2024pipnet} refined this approach by introducing categorical depth and multi-camera inputs, effectively predicting pedestrian intentions up to four seconds in advance on the PIE dataset.
Graph Convolutional Network (GCN) emerged as a compelling alternative, particularly for modeling spatial relationships within pedestrian trajectories. Unlike CNN and RNN, GCN excel in preserving the structural properties of pedestrian movement. \cite{8917118} pioneered a GCN-based approach that extracted pose-based skeletal features, while \cite{9833442} introduced a spatial-temporal GCN to capture dynamic motion dependencies. \cite{10421893} further refined this method with spatial, temporal, and channel attention mechanisms. However, pure GCN-based models often lacked environmental and vehicle motion awareness, prompting \cite{9774877} to integrate pedestrian image crops, skeleton data, speed, and segmentation maps with an attention mechanism for improved feature fusion.\\
The rise of Transformers introduced significant improvements in long-range dependency modeling. \cite{10247098} developed a Transformer framework incorporating pedestrian image crops, skeletal features, global scene images, bounding boxes, and ego-vehicle velocity. By leveraging self-attention and temporal fusion blocks, this model effectively captured dynamic spatiotemporal interactions. \cite{Zhang_Tian_Ding_2023} addressed AI uncertainty in challenging scenarios by integrating bounding box features with deep evidential learning. Hybrid fusion-based models have demonstrated superior predictive performance by integrating multiple feature extraction techniques \cite{ham2022mcip}. \cite{10418196} combined GCN and RNN, utilizing GRU for processing bounding boxes and vehicle speed while applying GCN to skeleton data. CIPF (Crossing Intention Prediction using Feature Fusion Modules) \cite{Ham_2023_CVPR} introduced three specialized fusion modules, observational, contextual, and convolutional, leveraging RNN layers and attention mechanisms to process eight diverse pedestrian and vehicle input features. This model set a new benchmark on the PIE dataset, showcasing significant improvements in predictive accuracy. Complementarily, the Dual-STGAT architecture \cite{lian2025dual} addresses real-time considerations and efficient feature fusion. Its Pedestrian Module adopts a spatio-temporal graph attention network for movement cues, while a Scene Module aggregates visual, semantic, and motion inputs, employing self-attention and Efficient Channel Attention (ECA) to reduce redundancy. Evaluations on the PIE dataset confirm its capacity for high accuracy with minimal latency, highlighting the practicality of graph-based fusion strategies for real-world autonomous driving.

\subsection{Crossing Intention Prediction at Intersections}
Predicting pedestrian crossing intentions at intersections is essential for proactively preventing accidents and improving overall traffic safety. Recent research has primarily focused on off-board sensing methods due to their broader field of view and stable observation capabilities, offering significant advantages over vehicle-mounted (on-board) systems. Off-board systems typically employ stationary cameras, such as CCTV cameras to monitor pedestrian movements widely \cite{abdelrahman2025video}. These approaches leverage computer vision and machine learning techniques to analyze various visual and motion-related features, including pedestrian posture, gait, head orientation, trajectory data, and interactions with vehicles and infrastructure \cite{zhang2020bprediction, zhou2022prediction, zhao2019probabilistic}.
Several studies have demonstrated the efficacy of video-based analysis combined with RNN model for crossing prediction tasks. For instance, \cite{zhang2020bprediction} employed stationary surveillance cameras and applied computer vision tracking in combination with an Long Short-Term Memory (LSTM) neural network, achieving better accuracy in predicting pedestrian intentions at signalized crosswalks. Similarly, \cite{xiong2024research} integrated pedestrian-vehicle interaction data from cameras and LiDAR sensors at unsignalized intersections. They extracted key temporal and spatial features, such as pedestrian speed, vehicle distance, and time-to-collision, to train an LSTM-based model, which yielded high predictive accuracy. To enhance predictive performance further, \cite{zhou2022prediction} introduced a cross-stacked GRU network designed explicitly for off-board surveillance video. Their model fused multiple contextual inputs, including pedestrian posture, local environmental cues, and global contextual information, addressing limitations inherent to vehicle-mounted camera systems and providing robust predictions of pedestrian crossing behavior.\\
Beyond RNN-based methods, others have explored probabilistic and graphical models to capture the complexity of pedestrian crossing dynamics. \cite{zhao2019probabilistic} implemented a roadside LiDAR-based predictive system, utilizing a probabilistic Naïve Bayes variant to forecast pedestrian crossing intentions at unsignalized intersections and mid-block crossings. Their approach provided reliable predictions within a critical 0.5–3-second window, proving particularly effective in proactive pedestrian safety interventions. \cite{zhou2024pedestrian} utilized a Hidden Markov Model, integrating social interaction forces with high-dimensional pedestrian features such as posture, speed, and spatial relationships. This approach achieved higher prediction accuracy compared to conventional classifiers like Bi-LSTM, SVM, and Random Forest.
Graph-based modeling techniques have also emerged as promising methods for pedestrian intention prediction. \cite{zhou2023pedestrian} proposed a graph-centric learning framework leveraging surveillance video data. Their model involved constructing a pedestrian-centric environment graph consisting of appearance and motion sub-graphs, processed through GCN-based model. Combined with pedestrian-specific trajectory and skeletal features analyzed via an LSTM decoder, this integrated approach effectively captured spatiotemporal relationships, achieving state-of-the-art performance in early pedestrian crossing prediction tasks. \cite{zhang2021pedestrian} utilized pose estimation model to extract more information about pedestrian body movements by analyzing CCTV-derived trajectories and pose estimation (keypoint detection). Machine learning models to forecast crossing intentions effectively, highlighting the optimal prediction performance within a horizon of 2 seconds. To support diverse pedestrian safety needs, \cite{yang2022cooperative} proposed the Vision Enhanced Nonmotorized Users Services (VENUS), an intelligent infrastructure node integrating real-time sensing, computer vision, and edge artificial intelligence. VENUS provides directional crossing requests and mobility information for pedestrians and cyclists, including persons with disabilities, thereby proactively addressing safety and accessibility simultaneously.

%% file: sec/3_CIPI.tex
\section{Methodology}
\label{sec:CIPI}
\subsection{VRU Crossing Prediction and Crossing Monitoring Framework}
\begin{figure*}
    \centering
    \includegraphics[width=\linewidth]{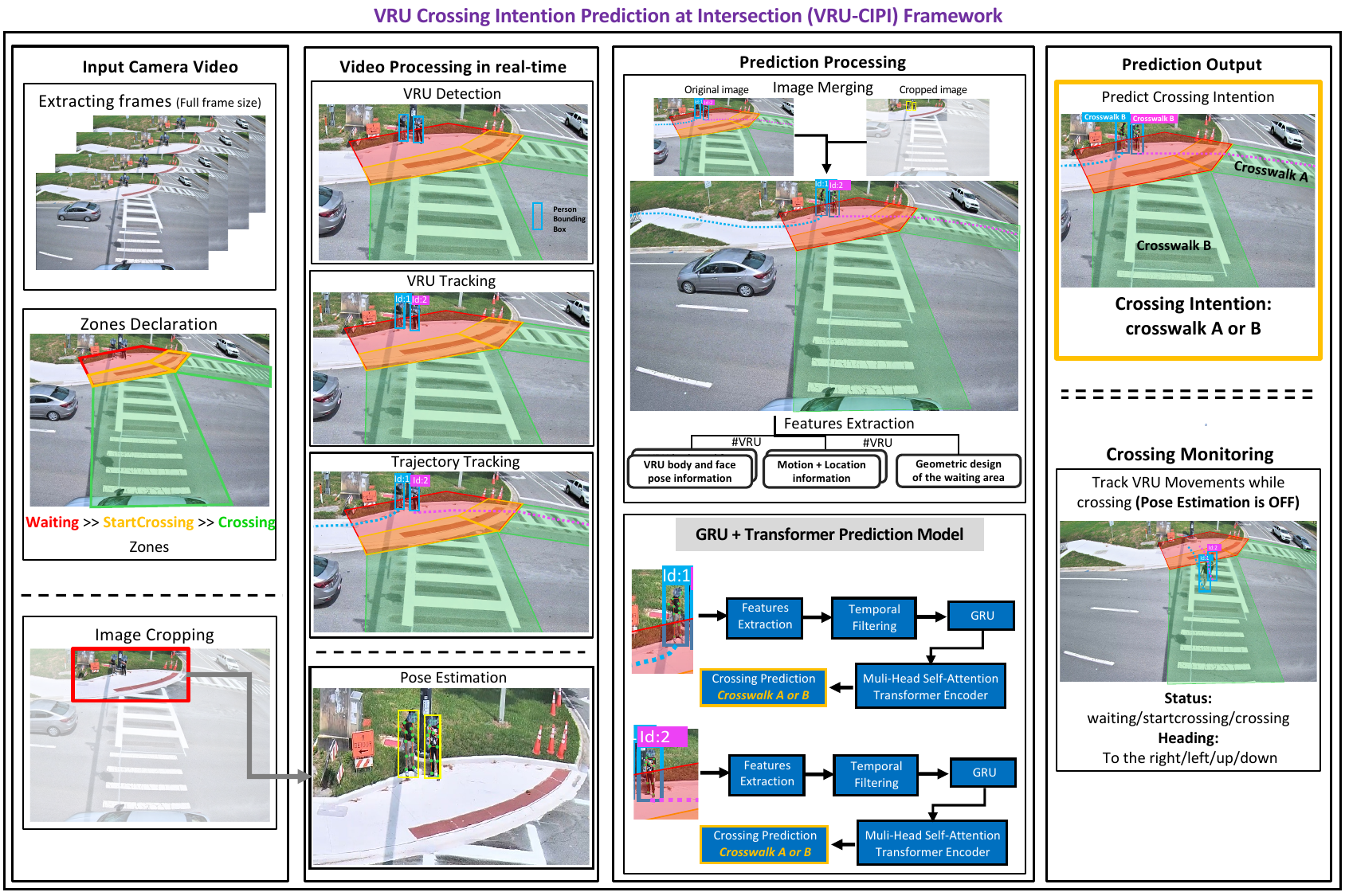}
    \caption{\textbf{VRU-CIPI framework for VRU monitoring and crossing prediction at intersections.} It integrates detection, tracking, and pose estimation models to extract individual features while considering the geometric design of the waiting area for accurate prediction.}
    \label{fig:CIPI_fw}
\end{figure*}

The proposed framework for monitoring and predicting the crossing intentions of VRUs consists of four main stages as visualized in Figure~\ref{fig:CIPI_fw} starting from the input video stream from cameras to crossing prediction of VRUs at intersections as detailed below.

\subsubsection{First Stage | ZoI Declaration}
Frames are extracted sequentially from the streamed video, and specific zones of interest (ZoI) are defined. These zones include the waiting area, the start-crossing zone, and the crossing zone. The start-crossing zone is introduced because some VRUs begin moving toward the crosswalk before the pedestrian signal is activated. However, their presence in this zone indicates a high likelihood of crossing, regardless of the traffic signal status. By tracking VRU trajectories, their positions within these defined zones can be identified \cite{jeon2023leveraging}. A key feature extracted for each VRU is body and face pose, which is determined using a pose estimation model. However, some cameras are positioned far from the waiting area, reducing the accuracy of pose estimation. To address this issue, the waiting area region is cropped from the full image and then processed using the pose estimation model to extract VRU body keypoints.

\subsubsection{Second Stage | Video Processing and Feature Extraction}
The primary goal is to accurately predict the crossing direction of VRUs by detecting, tracking, and estimating their poses at intersections. YOLOv8m and YOLOv8m-pose \cite{yolov8_ultralytics} are employed for both VRU detection and pose estimation, respectively. VRU movements are tracked frame-by-frame using OC-SORT \cite{cao2023observation}, enabling the calculation of their walking speed and heading direction using positional information extracted across multiple frames. Additionally, pose estimation identifies keypoints of VRUs and calculates angles and distances between body joints, including the body pose angle, face pose angle, and the distance between the shoulder points. Since the geometric features of waiting areas remain constant within each area, these spatial attributes complement the motion and pose features. Distinctly separated waiting areas for each crosswalk simplify crossing predictions, whereas small waiting areas serving multiple crosswalks complicate predictions because a VRU’s position alone does not clearly indicate their intended direction. Consequently, accurate pose estimation features become critical, especially in constrained waiting areas. The extracted VRU features, walking speed, heading direction, and pose angles are integrated to robustly determine crossing intentions, with particular emphasis on pose estimation in compact waiting zones.

\subsubsection{Third Stage | Crossing Prediction}
Since the detection and tracking models are applied to the full image while the pose estimation model operates on the cropped image, merging these outputs is necessary to assign all extracted features to the correct person ID. The two images are combined, and the pose estimation data is linked to the original person ID by matching locations in the merged image. Once merged, motion and pose estimation features are assigned to each person ID. Features are extracted every 10 frames, corresponding to a 0.5-second time interval. The input window size is 2.5 seconds (50 frames), with a step size of 0.5 seconds (10 frames) for accurate predictions as analyzed by \cite{sharma2025predicting}. To ensure smooth readings, features extracted from every 10-frame segment are subjected to temporal filtering through averaging. The filtered features are then grouped into sequences of five and passed to the VRU-CIPI model. The VRU-CIPI model is based on GRU with multi-head self-attention Transformer, to predict the crossing behavior of each VRU.

\subsubsection{Fourth Stage | Crossing Monitoring}
The final stage outputs the crossing predictions for each VRU while continuously monitoring their movements. This proactive method enables the activation of the appropriate signal phase without depending on potentially inaccurate push-button activations. It also helps reduce violations by VRUs who cross without pressing the push button. Once a VRU enters the crosswalk zone, detection and tracking continue to monitor their crossing. The pose estimation model is deactivated at this stage, as it is only required for crossing prediction. This approach ensures efficient real-time processing of the entire framework.

\subsection{VRU-CIPI: VRU Crossing Intention Prediction at Intersections}
\begin{figure*}
    \centering
    \includegraphics[width=\linewidth]{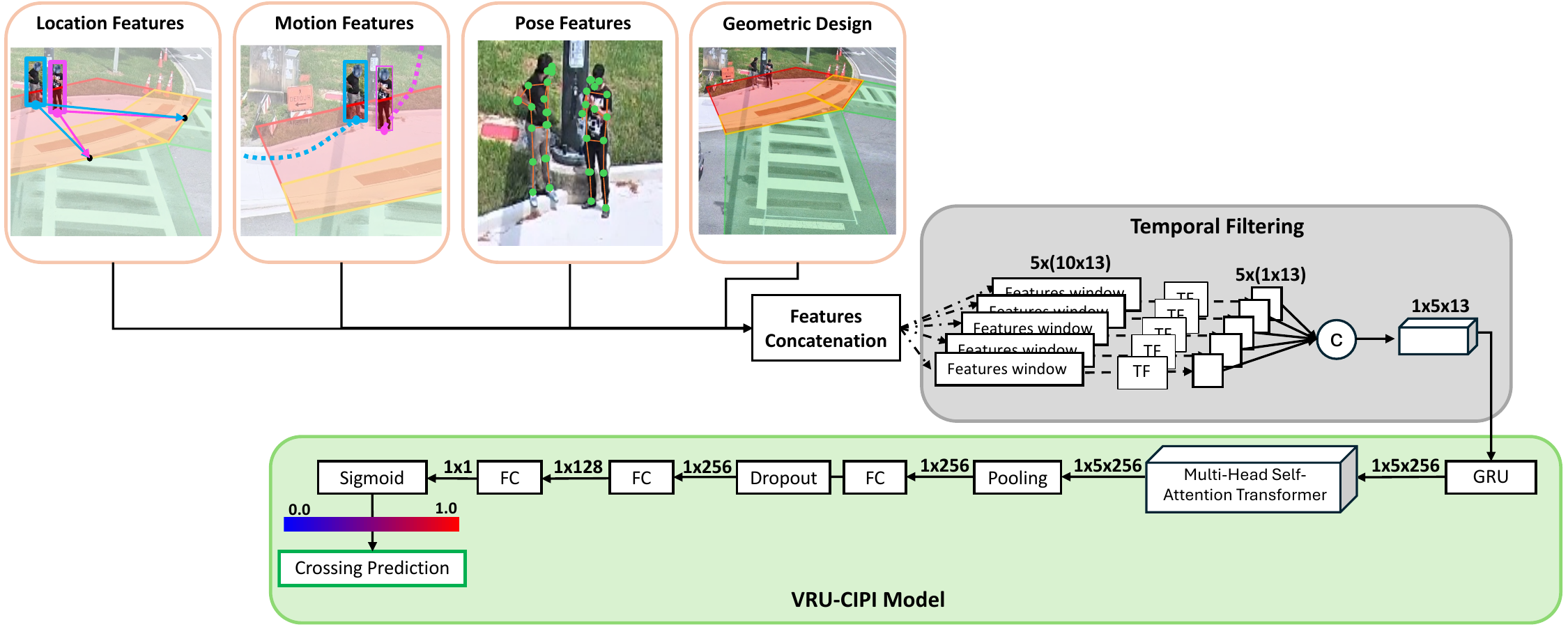}
    \caption{\textbf{VRU-CIPI Model Architecture for Crossing Intention Prediction.} Features are filtered, concatenated, and then passed to GRU to capture temporal dependencies, while multi-head self-attention transformer encoder models contextual relationships. The final prediction is made through fully connected layers with a sigmoid activation function. TF: temporal filtering. C: concatenation. FC: Fully Connected Layer.}
    \label{fig:vru_cipi}
\end{figure*}

Figure~\ref{fig:vru_cipi} shows the VRU-CIPI model architecture for crossing prediction. The model processes features through filtering and concatenation before sequential and contextual analysis is performed using a GRU \cite{chung2014empirical} and a multi-head self-attention transformer encoder \cite{vaswani2017attention, reza2022multi}. The GRU captures temporal dependencies within the input sequence $X = \{x_1, x_2, \ldots, x_T\}, \quad \text{where} \quad x_t \in \mathbb{R}^d$, by updating its hidden state \( h_t \) at each time step. The GRU uses an update gate \( z_t \) and a reset gate \( r_t \) to regulate the flow of information:

\begin{equation}
z_t = \sigma\left(W_z x_t + U_z h_{t-1} + b_z\right)
\end{equation}

\begin{equation}
r_t = \sigma\left(W_r x_t + U_r h_{t-1} + b_r\right)
\end{equation}

where \( W_z, W_r, U_z, U_r \) are learnable weight matrices, \( b_z, b_r \) are bias terms, and \( \sigma \) is the sigmoid activation function. The reset gate \( r_t \) modulates the influence of previous states on the candidate hidden state, defined as:

\begin{equation}
\widetilde{h_t} = \tanh{\left(W_h x_t + U_h \left(r_t \odot h_{t-1}\right) + b_h\right)}
\end{equation}

The final hidden state \( h_t \) is updated by balancing the contributions of the previous hidden state and the candidate hidden state through the update gate:

\begin{equation}
h_t = (1 - z_t) \odot h_{t-1} + z_t \odot \widetilde{h_t}
\end{equation}

The sequence of hidden states $H = \{h_1, h_2, \ldots, h_T\}$ produced by the GRU is forwarded to a Transformer encoder for contextual feature extraction. The Transformer encoder uses a scaled dot-product attention mechanism to capture relationships across the sequence. For the GRU outputs, the query (Q), key (K), and value (V) matrices are computed as:

\begin{equation}
Q = H W_Q, \quad K = H W_K, \quad V = H W_V
\end{equation}

where \( W_Q, W_K, W_V \) are learnable weights. The attention scores are then calculated as:

\begin{equation}
\text{Attention}(Q,K,V) = \text{softmax} \left(\frac{QK^T}{\sqrt{d_k}}\right) V
\end{equation}

The encoder of the VRU-CIPI model uses two attention heads to enhance its ability to capture diverse dependencies. The outputs of the two heads are concatenated and transformed using \( W_O \), a projection matrix to combine the heads, as follows:

\begin{equation}
\text{MultiHead}(Q,K,V) = \text{Concat}(\text{head}_1, \text{head}_2) W_O
\end{equation}

The output from the transformer encoder is passed through fully connected layers and a sigmoid activation function to predict VRU crossing behavior. By linking the Multi-Head Self-Attention outputs to the GRU, the architecture captures both contextual relationships across features and temporal dependencies within the sequence.

%% file: sec/5_Experiment.tex
\section{Experiemnts}
\label{sec:Exp}

\subsection{UCF-VRU Dataset}
This research focused on two major four-leg intersections in Florida, United States, with significant VRUs activity to develop a comprehensive dataset. These locations were chosen due to their proximity to prominent shopping areas and educational institutions, ensuring representation of a diverse spectrum of VRU categories. At each intersection, the study aimed to identify optimal camera positions to capture VRU movements at waiting areas and crosswalks comprehensively. Major intersections with a significant number of VRUs were specifically selected to evaluate the reliability of the crossing prediction model and its potential scalability to other locations. A total of 270 hours of video footage was collected using eight cameras (four per intersection), capturing diverse VRU activities under varying conditions, including daytime and nighttime, as well as clear and rainy weather. This dataset includes 962 VRUs categorized into pedestrians, non-motorized (Non-Mot) VRUs such as cyclists and scooter riders, and electric mobility (E-Mob) VRUs such as e-scooters, and e-wheelchairs. Nighttime data was particularly crucial for evaluating the model’s performance during periods of higher pedestrian accident risks \cite{lutkevich2012fhwa, schwendinger2023lighting}. Table~\ref{tab:ucf_vru_summary} provides a summary of the data collection, highlighting variations across different times of the day and weather conditions.

\begin{table}[h]
    \centering
    \caption{\textbf{Summary of UCF-VRU dataset.} It includes three primary VRU categories and under various conditions, including nighttime and rainy weather scenarios.}
    \resizebox{0.48\textwidth}{!}{%
    \begin{tabular}{l|ccc|cc|cc}
        & Ped & Non-Mot & E-Mob & Day & Night & Clear & Rain \\ 
        \midrule
        Inter. 1 & 332 & 156 & 70 & 497 & 59 & 511 & 45 \\ 
        Inter. 2 & 241 & 144 & 19 & 346 & 60 & 350 & 56 \\ 
        \midrule
        Total & 573 & 300 & 89 & 845 & 119 & 861 & 101 \\[1ex]
         & 59.6\% & 31.2\% & 9.2\% & 87.6\% & 12.4\% & 89.5\% & 10.5\% \\ 
    \end{tabular}}
    \label{tab:ucf_vru_summary}
\end{table}

\subsection{Evaluation Metrics}

To assess and compare the performance of the proposed model in predicting whether a VRU intends to cross at Crosswalk A or Crosswalk B, we utilized five widely adopted metrics in the pedestrian crossing intention prediction literature: accuracy (ACC), precision, recall, and F1-score. Accuracy measures how accurately the model predicts the binary classification problem of crossing intention, defined as follows,
\begin{equation}
    \text{ACC} = \frac{\text{TP} + \text{TN}}{\text{TP} + \text{TN} + \text{FP} + \text{FN}},
\end{equation}
where TP represents the quantity of true positive samples, TN represents the quantity of true negative samples, FP represents the quantity of false positive samples, and FN represents the quantity of false negative samples, respectively.

Precision measures the proportion of correct positive predictions among all positive predictions, defined as follows,
\begin{equation}
    \text{Precision} = \frac{\text{TP}}{\text{TP} + \text{FP}}.
\end{equation}

Recall measures the proportion of correct positive predictions among all actual positive cases, defined as follows,
\begin{equation}
    \text{Recall} = \frac{\text{TP}}{\text{TP} + \text{FN}}.
\end{equation}

F1-score is the harmonic mean of precision and recall, defined as follows,
\begin{equation}
    \text{F1} = \frac{2 \times \text{Precision} \times \text{Recall}}{\text{Precision} + \text{Recall}}.
\end{equation}

\subsection{Experimental Setting}
VRU-CIPI model was implemented on Nvidia RTX 3080 Ti and AMD Ryzen threadripper pro 3955wx 16-cores processor with pytorch environment. VRU-CIPI was trained on the UCF-VRU dataset for 250 epochs using the AdamW optimizer with an initial learning rate of \(2.5 \times 10^{-4}\). The learning rate was dynamically adjusted using a \textit{ReduceLROnPlateau} scheduler, reducing it by half whenever validation loss plateaued for two consecutive epochs. The model architecture consists of a GRU with 256 hidden units and two layers, integrated with a multi-head self-attention transformer encoder for capturing contextual relationships. Dropout was applied at a rate of 0.5 within the GRU, Transformer layers, and fully connected layers to reduce overfitting. Additionally, L2 regularization (weight decay) of \(1 \times 10^{-4}\) was utilized. To ensure stable gradients and improve training convergence, gradient clipping was employed with a maximum gradient norm of 1.0.

\subsection{VRU-CIPI Results}

The performance of the proposed VRU-CIPI model was comprehensively evaluated using the UCF-VRU dataset, a diverse dataset with various types of VRUs, including pedestrians, non-motorized VRUs such as cyclists, and electric mobility users like e-bikes and scooters, collected under daytime and nighttime as well as clear and rainy weather conditions. This variety ensures the robustness and generalizability of the model across real-world intersection scenarios. To the best of our knowledge, no prior relevant datasets captured from CCTV camera views include nighttime and rainy conditions. Table~\ref{tab:model_comparison} compares the performance of VRU-CIPI models with different numbers of attention heads. The proposed two-head VRU-CIPI achieved the highest accuracy (96.45\%), precision (96.38\%), recall (96.68\%), and F1-score (96.53\%). The single-head model exhibited slightly lower accuracy (96.27\%, a decrease of 0.18\%) but higher recall (97.63\%, an increase of 0.95\%). The four-head model attained an accuracy of 96.28\% (0.17\% lower than the two-head), with slightly reduced performance in other metrics. These results indicate that the two-head attention model optimally balances complexity and prediction accuracy. The entire framework, from reading video streams to crossing prediction, achieves real-time processing with an inference speed of 33 frames per second (FPS).

\begin{table}[ht]
    \centering
    \caption{\textbf{Performance of VRU-CIPI on UCF-VRU dataset.} Comparison of prediction results across different numbers of attention heads: single-head (VRU-CIPI-1), two-head (VRU-CIPI-2), and four-head (VRU-CIPI-4)}
    \resizebox{0.45\textwidth}{!}{%
    \begin{tabular}{lcccccc}
        \midrule
        \textbf{Model} & \textbf{Accuracy} & \textbf{Precision} & \textbf{Recall} & \textbf{F1-Score} \\ 
        \midrule
        VRU-CIPI-1 & 0.9627 & 0.9518 & \textbf{0.9763} & 0.9639  \\[1ex]
        
        VRU-CIPI-2 & \textbf{0.9645} & \textbf{0.9638} & 0.9668 & \textbf{0.9653} \\[1ex]
        
        VRU-CIPI-4 & 0.9628 & 0.9592 & 0.9682 & 0.9637 \\ 
        \bottomrule
    \end{tabular}}
    \label{tab:model_comparison}
\end{table}

\begin{figure}
    \centering
    \includegraphics[width=\linewidth]{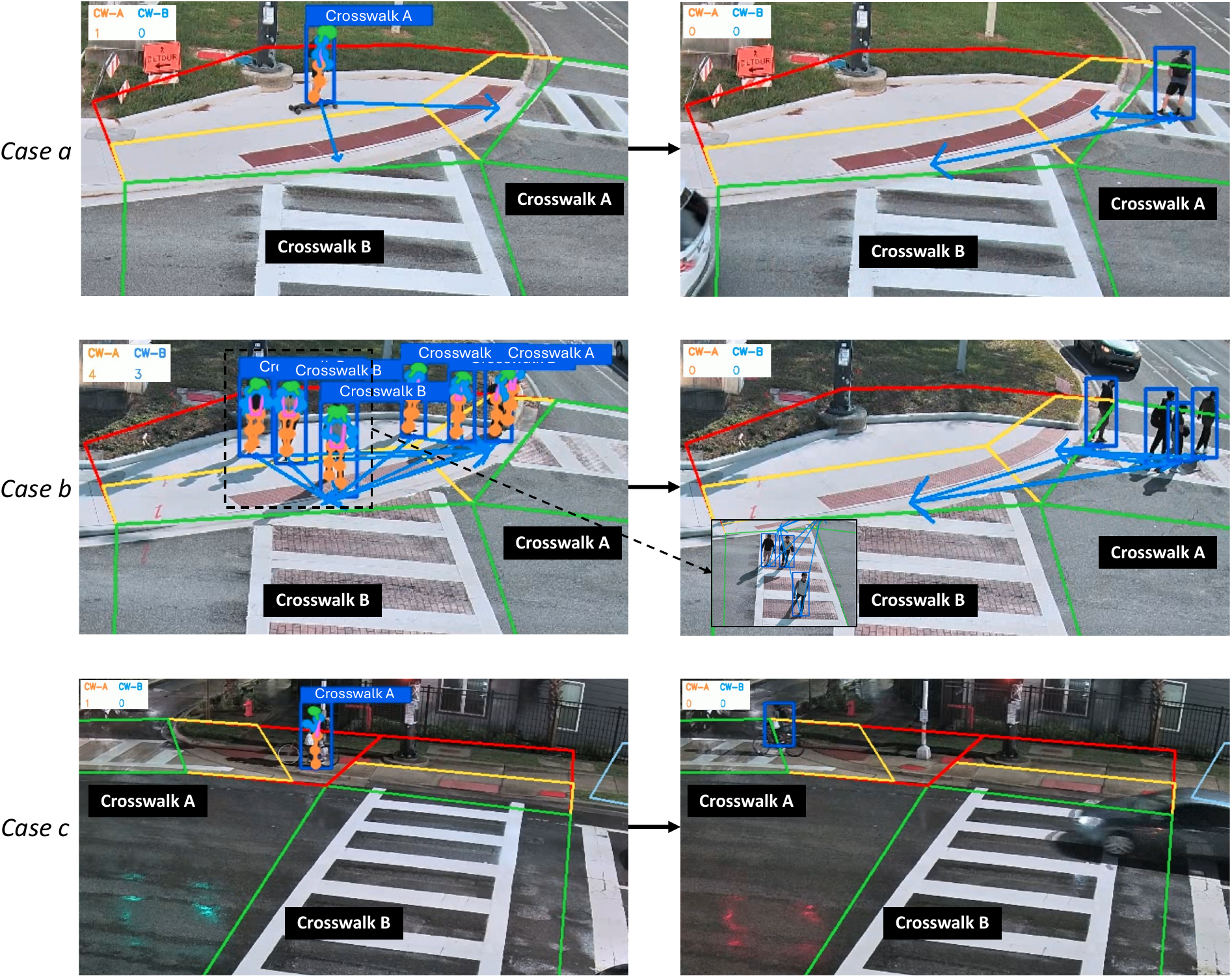}
    \caption{\textbf{The qualitative results of VRU-CIPI.} Examples illustrating model predictions for different VRU types and under different weather and lighting conditions. Cases include: (a) a scooter rider during clear daytime conditions; (b) a scenario with seven pedestrians; and (c) a cyclist during nighttime with rainy weather.}
    \label{fig:right_case}
\end{figure}

Figure~\ref{fig:right_case} presents qualitative results for VRU crossing intention prediction at intersections. Each case shows the crossing predictions of each VRU while waiting and after crossing, where the number of crossings per direction is visualized in the top left box. \textbf{Case a}, presents the crossing prediction of a VRU riding a skateboard where while the VRU based on the location features the VRU is nearer to crosswalk B, however, by analyzing other features, especially pose features, the prediction is correctly identified to be Crosswalk A. In \textbf{Case b}, several pedestrians are present with different crossing intentions, where VRU-CIPI is applied for predicting the crossing prediction for each VRU based on each aggregated features. \textbf{Case c} shows the crossing prediction of a cyclist at night and while it is raining, where the crossing prediction during such adverse weather conditions is crucial since the visibility of drivers decreases during such conditions. Such predictions enable timely warnings to vehicles, significantly improving VRU safety at intersections and other road junctions.

When compared to previous research (Table~\ref{tab:vru_comparison}), the proposed two-head attention VRU-CIPI model demonstrates superior performance across three defined categories. The comparison table is structured as follows: the first section presents models predicting VRU crossing intentions from a vehicle viewpoint, with the highest accuracy of 85\% achieved by PIP-Net \cite{azarmi2024pipnet}, where results reported are on PIE dataset and time-to-event is 2 seconds. The second section includes studies employing CCTV camera views to predict cross/no-cross behaviors, where the highest reported accuracy is 95.05\% by \cite{xiong2024research}, however, their dataset size is only includes 250 pedestrians which is relatively small. Finally, the third section is also from CCTV camera view but on models predicting crossing intentions for multiple crosswalks, which are particularly important at intersections. In this critical area, the proposed VRU-CIPI model achieves an accuracy of 96.45\%, substantially surpassing the previous best model by \cite{yang2022cooperative}, which reported 90.24\%. Furthermore, VRU-CIPI framework maintains real-time inference speed (33 FPS), unlike previous frameworks. \cite{yang2022cooperative} achieved only 2 FPS for their entire framework, and while \cite{zhou2023pedestrian} claimed 32 FPS for the prediction model alone, they stated that their complete framework could not sustain real-time processing. \cite{lian2025dual} reported a 5.01 ms latency for prediction model inference only, whereas our model achieves a significantly lower inference latency of 0.78 ms.\\
Utilizing CCTV cameras to predict VRU crossing intentions offers distinct advantages in intersection scenarios, providing more accurate, reliable, and comprehensive observations. Furthermore, coupling these predictions with (Vehicles-to-Everything) V2X and I2V communication technologies enables vehicles to receive precise, real-time alerts about the crossing direction of VRUs, significantly improving intersection safety for all road users.\\
It is important to recognize the limitations when directly comparing our framework to others. Due to differences in datasets and the unavailability of publicly shared models of CCTV view models, the evaluation focuses primarily on overall prediction model performance and dataset diversity. Additionally, our approach relies on fixed-camera setups and known intersection layouts to identify waiting areas, making it less generalizable compared to methods using vehicle-mounted cameras.



\begin{table}[ht]
    \centering
    \caption{\textbf{Comparison with existing person crossing intention prediction.} For each model, the visual encoder and input features are introduced, and their performance is compared. GAN: Graph Attention Network. GCN: Graph Convolutional Network.}
    \resizebox{0.48\textwidth}{!}{%
    \begin{tabular}{lcccccc}
        \midrule
        \textbf{Reference} & \textbf{Year} & \textbf{Model Type} & \textbf{VRU Types} & \textbf{Accuracy} \\ 

        \midrule 
        MCIP \cite{ham2022mcip} & 2022 & GRU+Attention & 1 & 83\% \\[1ex] 
        
        CIPF \cite{Ham_2023_CVPR} & 2023 & GRU+Attention & 1 & 84\% \\[1ex] 

        PIP-Net \cite{azarmi2024pipnet} & 2024 & GRU+Attention & 1 & 85\% \\[1ex] 

        Dual-STGAT \cite{lian2025dual} & 2025 & GAN+LSTM & 1 & 83\% \\[1ex] 
        
        \midrule

        Zhang et al. \cite{zhang2020bprediction} & 2020 & LSTM & 1 & 91.60\% \\[1ex]

        Liang et al. \cite{liang2023predicting} & 2023 & CNN & 2 & 84.96\% \\[1ex]

        Zhou et al. \cite{zhou2023pedestrian} & 2023 & GCN+LSTM & 1 & 75.7\% \\[1ex] 

        Xiong et al. \cite{xiong2024research} & 2024 & LSTM & 1 & 95.05\% \\[1ex]
        
        \midrule

        VENUS \cite{yang2022cooperative} & 2022 & CNN & 6 & 90.24\% \\[1ex]
        
        \textbf{VRU-CIPI (ours)} & 2025 & GRU+Transformer & 5 & \textbf{96.45\%} \\
        \bottomrule
    \end{tabular}}
    \label{tab:vru_comparison}
\end{table}

\subsection{Ablation Study}

An ablation study was conducted to evaluate the incremental contribution of four feature sets Location (L), Motion (M), Geometric design (G), and Pose estimation (P) to the VRU-CIPI model performance (Table~\ref{tab:cipi_features_comparison}). Starting with location features alone, the model achieved a baseline accuracy of 92.72\%. Adding motion features resulted in a modest accuracy increase of 0.26\% (92.98\%) and improved recall (+1.09\%), highlighting better temporal capture of VRU dynamics. Incorporating geometric design features further boosted accuracy by 0.30\% (93.28\%) and significantly increased precision (+1.16\%), enhancing prediction in complex intersections. Most notably, adding pose estimation features led to a substantial accuracy improvement of 3.17\%, reaching the highest accuracy (96.45\%), precision (96.38\%), recall (96.68\%), and F1-score (96.53\%). This significant gain underscores the critical role of pose estimation features such as body orientation and face looking angle in accurately identifying crossing intentions, particularly when position alone is ambiguous.

\begin{table}[ht]
    \centering
    \caption{\textbf{Ablation study of the VRU-CIPI model with incremental feature combinations.} Incorporating all features, particularly pose estimation, significantly improves prediction accuracy.  Location (L), Motion (M), Geometric design (G), and Pose estimation (P).}
    \resizebox{0.48\textwidth}{!}{%
    \begin{tabular}{cccccccc}
        \midrule
        \multicolumn{4}{c}{\textbf{Features}} & \textbf{Accuracy} & \textbf{Precision} & \textbf{Recall} & \textbf{F1-Score} \\ 
        \cmidrule(lr){1-4}
        L & M & G & P & & & & \\ 
        \midrule
        $\checkmark$ & & & & 0.9272 & 0.9379 & 0.9179 & 0.9278  \\[1ex]
        
        $\checkmark$ & $\checkmark$ & & & 0.9298 & 0.9333 & 0.9288 & 0.9310 \\[1ex]
        
        $\checkmark$ & $\checkmark$ & $\checkmark$ & & 0.9328 & 0.9449 & 0.9291 & 0.9369 \\[1ex]
        
        $\checkmark$ & $\checkmark$ & $\checkmark$ & $\checkmark$ & \textbf{0.9645} & \textbf{0.9638} & \textbf{0.9668} & \textbf{0.9653} \\ 
        \bottomrule
    \end{tabular}}
    \label{tab:cipi_features_comparison}
\end{table}

%% file: sec/6_Conc.tex
\section{Conclusion}
\label{sec:conc}

Analyzing human behavior to predict their crossing intentions at urban intersections, remains crucial for enhancing interaction safety between road users. In this paper, we introduced VRU-CIPI, a prediction framework based on GRU with multi-head self-attention, designed to predict the crossing intentions of VRUs including pedestrians, cyclists, and other VRU categories at intersections. The proposed model leverages various input features including location, motion trajectories, human pose estimations, and geometric attributes of intersection waiting areas. We gathered the UCF-VRU dataset, capturing diverse VRU behaviors under different conditions, including daytime and nighttime, as well as clear and rainy weather scenarios. VRU-CIPI achieved state-of-the-art crossing prediction accuracy of 96.45\%, operating effectively in real-time at 33 FPS for the whole framework. By integrating the proposed method with V2X and I2V communication to connected vehicles, our approach proactively improves intersection safety by promptly activating pedestrian crossing signals while simultaneously supporting autonomous driving through early warnings of the crossing events.

